\documentclass[letterpaper]{article} % DO NOT CHANGE THIS
\usepackage[submission]{aaai23}  % DO NOT CHANGE THIS
\usepackage{times}  % DO NOT CHANGE THIS
\usepackage{helvet}  % DO NOT CHANGE THIS
\usepackage{courier}  % DO NOT CHANGE THIS
\usepackage[hyphens]{url}  % DO NOT CHANGE THIS
\usepackage{graphicx} % DO NOT CHANGE THIS
\usepackage{natbib}  % DO NOT CHANGE THIS AND DO NOT ADD ANY OPTIONS TO IT
\usepackage{caption} % DO NOT CHANGE THIS AND DO NOT ADD ANY OPTIONS TO IT
\usepackage{algorithm}
\usepackage{algorithmic}

\usepackage{soul}
\usepackage{color,xcolor}
\usepackage{makecell}
\usepackage{multirow}
\usepackage{threeparttable}
\usepackage{newfloat}
\usepackage{listings}
\DeclareCaptionStyle{ruled}{labelfont=normalfont,labelsep=colon,strut=off} % DO NOT CHANGE THIS
\floatstyle{ruled}
\newfloat{listing}{tb}{lst}{}
\floatname{listing}{Listing}
\title{Ultra-low Latency Adaptive Local Binary Spiking Neural Network with Accuracy Loss Estimator}
\author{
    Changqing Xu\textsuperscript{\rm 1}\thanks{cqxu@xidian.edu.cn.}\equalcontrib
    Yijian Pei\textsuperscript{\rm 1}\equalcontrib
    Zili Wu\textsuperscript{\rm 3},
    Yi Liu\textsuperscript{\rm 2},
    Yintang Yang\textsuperscript{\rm 2}\thanks{ytyang@xidian.edu.cn.}
}

\affiliations{
    \textsuperscript{\rm 1}Guangzhou Institute of Technology, Xidian University\\
    \textsuperscript{\rm 2} School of Microelectronics, Xidian University\\
    \textsuperscript{\rm 3} School of Computer Science and Technology, Xidian University
}

\usepackage{amsmath}
\usepackage{booktabs}

\begin{document}

\maketitle

\begin{abstract}
%Inspired by biological neurons, spiking neural networks(SNNs) has higher information processing ability and computational energy efficiency, so it is considered to be the third generation neural networks.
Spiking neural network (SNN) is a brain-inspired model which has more spatio-temporal information processing capacity and computational energy efficiency.
%However, with the deepening of the depth of SNNs, the memory problem caused by the weight of SNNs has gradually attracted attention.
However, with the increasing depth of SNNs, the memory problem caused by the weights of SNNs has gradually attracted attention.
%Inspired by convolutional neural networks(CNNs) quantization technology, there are two main methods to solve this problem: (1) binary convolution SNNs(BCSNNs) is obtained from trained binary convolution neural networks (BCNNs), (2) train binary convolutional SNNs directly.
%Inspired by Artificial Neural Networks (ANNs) quantization technology, there are two main methods to solve this problem. One is to obtain binary SNNs (BSNNs) from trained binary Artificial Neural Networks (BANNs), and the other is to obtain the binary SNN by training directly.
Inspired by Artificial Neural Networks (ANNs) quantization technology, binarized SNN (BSNN) is introduced to solve the memory problem.
Due to the lack of suitable learning algorithms, BSNN is usually obtained by ANN-to-SNN conversion, whose accuracy will be limited by the trained ANNs.
%However, the accuracy of the former is limited by the trained BCNNs, the accuracy of the latter is generally lower than the former due to the lack of a suitable training algorithm and the training time is expensive.
%However, the accuracy of the former will drop due to the conversion from BANN to BSNN, the accuracy of the latter is generally lower than the former due to the lack of a suitable training algorithm and the training time is expensive.
In this paper, we propose an ultra-low latency adaptive local binary spiking neural network (ALBSNN) with accuracy loss estimators, which dynamically selects the network layers to be binarized to ensure the accuracy of the network by evaluating the error caused by the binarized weights during the network learning process.
Experimental results show that this method can reduce storage space by more than 20 $\%$ without losing network accuracy.
At the same time, in order to accelerate the training speed of the network, the global average pooling(GAP) layer is introduced to replace the fully connected layers by the combination of convolution and pooling, so that SNNs can use a small number of time steps to obtain better recognition accuracy.
In the extreme case of using only one time step, we still can achieve 92.92 $\%$, 91.63 $\%$ ,and 63.54 $\%$ testing accuracy on three different datasets, Fashion-MNIST, CIFAR-10, and CIFAR-100, respectively.
%whose best classification accuracy of our network reached 92.42 $\% $, 90.12 $\% $and 63.54 $\%$on Fashion-MNIST, CIFAR-10, and CIFAR-100 datasets).
%\hl{It is a computational paradigm that simulates the biological brain based on the dynamic activation of binary neurons and event-driven?}
\end{abstract}

\section{Introduction}

Courbariaux et al. proposed BinaryConnect\cite{courbariaux2015binaryconnect}, which pioneered the study of binary neural networks. 
Binarization can not only minimize the storage usage and computational complexity of the model but also reduce the storage resource consumption of model deployment and greatly accelerate the inference process of the neural network.
In the field of CNNs, many algorithms have been proposed and satisfactory progress has been made.
Spiking neural networks, as the third generation of neural networks, is a computational paradigm that simulates the biological brain based on the dynamic activation of binary neurons and event-driven\cite{tavanaei2019deep,illing2019biologically}. 
By making use of the time sparsity of binary time series signals, it can improve the computational energy efficiency on special hardware\cite{mead1990neuromorphic}. 
The combination of SNNs and the binary networks has gradually attracted more and more attention\cite{lu2020exploring,kheradpisheh2022bs4nn,srinivasan2019restocnet}.  
However, it is still a great challenge to train SNNs due to their non-differentiable activation function. 
In order to maintain good accuracy, some researchers choose to use pre-training to obtain parameters from ANNs\cite{lu2020exploring,wang2020deep,cao2015spiking}. 
But the pre-training of ANNs gives up the advantage of SNNs in temporal and spatial information processing. 
In recent years, some studies have successfully trained binarized SNNs, directly. 
For example, Jang et al.\cite{jang2021bisnn} used the Bayesian rule to directly train binarized SNNs(BSNNs), and Kheradpisheh et al.\cite{kheradpisheh2022bs4nn} used time-to-first-spike coding in the direct training of the network.

We find that it is more reasonable to train BSNNs directly, but we need to build a reasonable SNNs structure and improve the learning algorithm appropriately. 
Therefore, we propose Accuracy Loss Estimator (ALE) and Global Average Pooling (GAP) Layer and use them to construct an ultra-low latency adaptive local binary spiking neural network.
We directly train the network using the iterative neuron model in \cite{wu2018spatio}, then the binarized layer is automatically selected by ALE to solve the problem of large precision loss in direct training.
Secondly, we use the GAP layer instead of the fully connected layer to reduce the amount of calculation and change the output layer of SNNs to alleviate the phenomenon that it takes a long time to train BSNNs directly.
Object recognition experiments are conducted on three different datasets: Fashion-MNIST, CIFAR-10, and CIFAR-100, and a comprehensive comparison is made with other BSNNs.
Through experiments, we verify the effectiveness of ALE and demonstrate the advantages of our method in terms of accuracy and training time.

\section{Related Works}
\subsection{Binary Spiking Neural Networks} 
Generally, when choosing the quantization of the network, we can consider the following two aspects: weight and input\cite{qin2020binary}. 
However, due to the characteristics of SNNs, there is no need to apply extra additional quantization of the network input. 
Recently, the idea of combining the SNN and the binarization has been proposed.
Lu et al.\cite{lu2020exploring} proposed B-SNN, which is transformed into BSNNs by pre-training BCNNs.
Roy et al.\cite{roy2019scaling} analyzed the results of combining different binary neurons with various binarized weight methods.
Kheradpisheh et al.\cite{kheradpisheh2022bs4nn} proposed BS4NN and explored the adaptation of simple non-leaky integrate-and-fire neurons, time-to-first-spike coding, and binarized weight in backpropagation.
Jang et al.\cite{jang2021bisnn} proposed BISNN, which combined Bayesian learning to train SNNs with binarized weights. 
However, a lot of work has focused on approximating full precision weights or reducing gradient errors to learn discrete parameters.
For BSNN, it is usually to keep the first layer and the last layer binarized to reduce the accuracy drop based on the experimental experience\cite{deng2021comprehensive}. 
This method usually works, but there is still room for improvement.

\subsection{Training of Binary Spiking Neural Networks}
The training methods of BSNNs are also getting more and more attention.
Recently, Mirsadeghi et al.\cite{MIRSADEGHI2021131} proposed the STiDi-BP algorithm to avoid reverse recursive gradient computation while using binarized weights to obtain good performance.
Wang et al.\cite{wang2020deep} proposed the weights-thresholds balance conversion method to scale the full precision weights into binarized weights through changing the corresponding thresholds of spiking neurons, then effectively obtain BSNNs.Roy et al.\cite{roy2019scaling}trained ANNs with
constrained weights and activations and deployed them into SNNs with binarized weights.
The BS4NN proposed by Kheradpisheh et al.\cite{kheradpisheh2022bs4nn} takes advantage of the temporal dimension and performs better than a simple BNNs with the same architecture.
The current BSNNs training method mainly uses all binarized weights, which fails to achieve a balance between accuracy and spatial quantization. 
Furthermore, SNNs usually require sufficient time steps to simulate neural dynamics and encode information, and also take a long time to converge, which brings huge computational costs\cite{sengupta2019going}.

\section{Approach}
In this section, we will introduce the neuron model, binary spiking neural network learning method, and binarization method first. Then we will also introduce our proposed accuracy loss estimator and GAP Layer.

\subsection{Iterative Leaky Integrate-and-Fire Neural Model}
In this paper, we use iterative Leaky Integrate-and-Fire(LIF) neuron model to construct networks.
First, we will introduce the classic Leaky Integrate-and-Fire (LIF) model, which is defined as:
\begin{equation}
\tau \frac{du(t)}{dt} = -u(t) + I(t),u < V_{th}
\label{eq1}
\end{equation}
where $u(t)$ is the membrane voltage of the neuron at time $t$, $\tau$ is the decay constant of the membrane potential, and $I(t)$ is the input from the presynaptic neuron. The membrane potential $u$ exceeds the threshold $V_{th}$ and then returns to the resting potential after firing a spike.

Then, the LIF neuron model is converted into an iterative version that is easy to program. Specifically, an iterative version can be obtained by the last spiking moment and the pre-synaptic input:
\begin{equation}
u(t_i) = u(t_{i-1})e^{\frac{t_{i-1}-t}{\tau}} + I^{'}(t_i)
\label{eq2}
\end{equation}
where $u(t_{i-1})$ is the membrane voltage at time step $t_{i-1}$ and the $I^{'}(t)$ is the input from the presynaptic neuron at time step $t_i$. 

When the neuron output is zero before the last moment, the membrane voltage begins to leak. This process can be expressed mathematically simply,
%$S = \tau (1-o^{t,n+1}(i))$

%\begin{equation}
%    u(t_i) = u(t_{i-1})\tau (1-o(t_i)) + I^{'}(t_i)
%%    u(t_{t+1})=\tau (1-o^{t,n+1}(i))u(t_{t})
%    \label{eq2_1}
%\end{equation}

%where $o(t_i)$ is the output of the neuron at time step $t_i$.
%To sum up, the iterative leaky integrate-and-fire(LIF) neuron model can be expressed as:
\begin{equation}
%u^{t+1,n+1}(i)= \tau u^{t,n+1}(i)(1-o^{t,n+1}(i))+\sum^{l(n)}_{j=1}w^n_{ij}o^{t+1,n}(j)
u^{l+1}_p(t_{i+1})= \tau u^{l+1}_p(t_i)(1-o^{l+1}_p(t_i))+\sum^{l_{max}}_{q=1}w_{pq}o^{l}_q(t_{i+1})
\label{eq3}
\end{equation}
where $u^{l+1}_p(t_{i+1})$ is the membrane voltage of $p$th neuron of $(l+1)$th layer at time step $t_{i+1}$, and $o^{l+1}_p(t_i)$ is the output of $p$th neuron of $(l+1)$th layer at time step $t_i$, $\tau$ is the decay factor, $w_{pq}$ represents the weight of the $q$th synapse to the $p$th neuron and $l_{max}$ is the total number of neurons at the $l$th layer. 

Finally, a step function $f(x)$ is used to represent whether the neuron's membrane voltage reaches a threshold voltage $V_{th}$ and fires a spike:
\begin{equation}
%o^{t+1,n+1}(i) = f(u^{t+1,n+1}(i))
o^{l+1}_p(t_{i+1}) = f(u^{l+1}_p(t_{i+1}))
\label{eq4}
\end{equation}
where the step function is $f(x)= \begin{cases} 1 & x \geq\ V_{th} \\ 0 & x < V_{th} \end{cases} $

\subsection{Accuracy loss estimator for weight binarization}
In order to reduce the accuracy drop of BSNNs, it is usually to keep the first and last layers non-binarized based on engineering experience, which means that the weight precision of the first and last layers plays a important role in the inference of the neural network\cite{deng2021comprehensive}.
However, according to our study, which layer should be binarized is dependent on the structure of the neural networks and characteristics of datasets, and it is not always the best solution to keep the first and last layers with full precision.
Therefore, we propose ALE, which automatically selects binarized and non-binarized network layers during network training by estimating the effect of different network layers on network accuracy.

First of all, we use the Manhattan Distance between approximate binarized weights and full precision weights as the error estimation of binarized weight $w_{loss}^l$, its calculation formula is shown below:
\begin{equation}
%w_{loss}^l = \sum^n_{i=1}|w_i^l-bin_{w_i^l}|,l =1,2,3...L
w_{loss}^l = \sum^n_{i=1}|w_i^l-bw_i^l|, l =1,2,3...L
\label{eq5}
\end{equation}
where $w^l_i$ is the $i$th full precision weight of the $l$th layer, $bw^l_i$ is the $i$th approximate weight of the $l$th layer

%However, this turn out to create additional issues later.
%On the one hand, we found that in an untraditional network structure, the Manhattan Distance of all weights cannot well represent the trend of accuracy change.
%On the other hand, $w_{loss}$ will increase rapidly as the number of weights increases, this is not conducive to calculation.

%But we can look at it from another perspective, the $w_{loss}$ of a convolution kernel is small, the ability of feature extraction is strong, so the result is good, and multiple output channels are equivalent to combining multiple good results.
%On the contrary, the $w_{loss}$ of a convolution kernel is large, the ability of feature extraction is weak, so the result is poor, and multiple output channels are equivalent to combining multiple poor results.
%Therefore, we use the $w_{loss}$ of the convolution kernel on a single channel to describe the error estimation of weights.
%As follows: \hl{**?}
For a BSNN, each output channel of the spiking convolution layer corresponds to one feature extraction. 
So we use the average error of feature extraction $A^l$ to estimate the error caused by the binarized weights. The formula is shown below. 
\begin{equation}
A^l=\frac{w^l_{loss}}{c_{out}^l}
\label{eq6}
\end{equation}
where $c_{out}^l$ is the number of output channels of the $l$th layer.

%Then, we consider the amount of storage space used by the weights of different layers. 
Besides the error caused by binarization, we also consider the size of weight storage space as the criteria for selecting binarized layers.
%For weights of each layer, a corresponding value $M$ needs to be selected to weigh the influence of the storage space on the selection result.
%We want a layer with a larger number of weights to be selected to maintain full accuracy with a lower probability, which will also lead to the result of preferring layers with a lower total number of weights.
We hope that the layer with a large number of weights will have a greater probability of being selected for binarization.
%Considering that the units of the two variables are different and difficult to combine, we try to use the weight error to estimate the difference in the weight storage space of different layers.
Because error estimation caused by binarization $A$ is calculated based on $w_{loss}$, we try to use $w_{loss}$ to estimate the difference in the weight storage space of different layers $M$, the formula is as follows:
\begin{equation}
M^l= \frac{\theta_{max}^l -\theta_1^l}{2} 
\label{eq7}
\end{equation}
$\theta_{max}^l$ is the $A^l$ obtained when the number of output channels of the $l$th layer is equal to 1, and $\theta_{1}^l$ is the obtained $A^l$ when the number of output channels of the $l$th layer equal to the total number of weights.
For example, for a weight in the shape of $[output\ channel,input\ channel,kernel\ size,kernel\ size]\\ =[10,10,3,3]$, its $\theta_{max}$ is equal to $A^l$ in the shape of $[1,100,3,3]$, and $\theta_{1}$ is equal to $A^l$ in the shape of $[100,1,3,3]$. These $A^l$ can be obtained quickly by using the equality \eqref{eq5} and \eqref{eq7}.

To simply the calculation of $M$, we use the $A^l$ to estimate the $\theta_{1}^l$ and $\theta_{max}^l$ based on the relationship between the error estimation of binarization weights with different shapes, which is obtained by experiments. The relationship is shown below.
%Before setting $M$, we observed that when a certain $w1_{loss}$ is known, the $w2_{loss} $ corresponding to any shape weight can be obtained by changing the number of input and output channels, as shown in equality\eqref{eq7}.
\begin{equation}
\frac{w^l_{loss}}{w^{l'}_{loss}} \approx \sqrt[2]{(\frac{c_{out}^l}{c_{out}^{l'}})^2 * \frac{c_{in}^l}{c_{in}^{l'}}}
\label{eq8}
\end{equation}
where $w_{loss}^l$, $c_{out}^l$, $c_{in}^l$ are the weight error of $l$th layer', the number of output channels, and the number of input channels, respectively. $w_{loss}^{l'}$, $c_{out}^{l'}$, $c_{in}^{l'}$ are the weight error of $l$th layers reshaped weights, the corresponding number of output channels, and the corresponding number of input channels, respectively.

%For the scale factor M that measures the size of the weight space, we consider to obtain it from two extreme cases of $w_{loss}$, the formula is as follows:
%\begin{equation}
%M^l= \frac{\theta_{max}^l -\theta_1^l}{2} 
%\label{eq8}
%\end{equation}
%$\theta_{max}^l$ is the $A^l$ obtained when the number of output channels of the $l$th layer is equal to 1, and $\theta_{1}^l$ is the obtained $A^l$ when the number of output channels of the $l$th layer equal to the total number of weights.
%For example, for a weight in the shape of $[output-channel,input-channel,kernel-size,kernel-size] = [10,10,3,3]$, its $\theta_{max}$ is equal to $A^l$ in the shape of $[1,100,3,3]$, and $\theta_{1}$ is equal to $A^l$ in the shape of $[100,1,3,3]$. These $A^l$ can be obtained quickly by using the equality \eqref{eq5} and \eqref{eq7}.

%Furthermore, we consider the influence of approximate weights at different layers in forward and back propagation.
%After experiments on the hierarchical structure, we found that the binarization operation on the first layer will have a greater impact on the accuracy. In addition, for the last layer, backpropagation needs to be performed to adjust the weights.
%As the starting first layer, the last layer will also have a greater impact on the accuracy. So, we directly set the first and last layers to have the same degree of influence on the result, and use a parabola to describe this phenomenon:
Furthermore, we consider the influence of binarized weights at different layers in the forward pass and backpropagation
After experiments on the hierarchical structure, we found that the binarization operation on the first layer will have a greater impact on inference accuracy. In addition, for backpropagation, the last layer has a greater impact on training. So, we directly set the first and last layers to have the same degree of influence on the result, and use a parabola $F(x)$ to describe this phenomenon:
\begin{equation}
F(x) = \epsilon(x - \frac{sumL+1}{2})^2
\label{eq9}
\end{equation}
where $x$ is the index of layer, $\epsilon$ is a facter which is equal to $\frac{4 * \eta}{sumL^2}$, $sumL$ represents the total number of layers, $\eta$ is a variable, and we set it to 1 by default. 

We combine $A^l$, $M^l$, and $F(x)$ together to get the criteria $R(x)$ for selecting binarized layers, which is shown below.
%We use $K$ to represent the number of classes in the dataset and act on the second half of the parabola.
%Here we summarize these three aspects to get the formula:
%\begin{equation} R(x) = \begin{cases}(\frac{1}{A^l+M^l})\epsilon(x - \frac{sumL+1}{2})^2&, x \leq \frac{sumL+1}{2} \\ (\frac{1}{A^l+M^l})log_{10}(K)\epsilon(x - \frac{sumL+1}{2})^2&, x > \frac{sumL+1}{2} \\\end{cases} 
%\label{eq10}
%\end{equation}

\begin{equation} R(x) = \begin{cases}(\frac{1}{A^l+M^l})F(x)&, x \leq \frac{sumL+1}{2} \\ (\frac{1}{A^l+M^l})log_{10}(K)F(x)&, x > \frac{sumL+1}{2} \\\end{cases} 
\label{eq10}
\end{equation}
where $K$ to represent the number of classes in the dataset.
%We can choose Standards according to the result R obtained by ALE, so that some layers keep full precision and some layers use binary weight. We will discuss the choice of Standards in the experiment section.
We can make different selection strategies according to the value of $R(x)$ to satisfy different applications.
We will discuss the strategies in detail in the experiment section.

%$$SNN \mathop{\rightarrow }\limits R(x) \mathop{\rightarrow }\limits^{strategy} BSNN$$

\subsection{Backpropagation with adaptive local binarization}

For the binarization of the weights, we use 3 binarized weight blocks for the binarization approximation of the full precision weights. That is, a linear combination of 3 binary filters $\alpha$ is used to represent the full precision weight $W$
\begin{equation}
W \approx \alpha_1B_1 + \alpha_2B_2 + \alpha_3B_3
\label{eq12}
\end{equation}

Then we calculate the value of each binarized weight $B$ referring to \cite{lin2017towards}. The equations are given as follows:
\begin{equation}
B_i = sign(W - mean(W) + (i-2)std(W)),i = 1,2,3
\label{eq13}
\end{equation}
where $mean(W)$ and $std(W)$ are the mean and standard deviation of $W$, respectively.

%Once $B$ is obtained, the equality\eqref{eq12} becomes a linear regression problem:
Once $B$ is obtained, we can get $\alpha$ easily according to:
\begin{equation}
\mathop{min}\limits_{\alpha}J(\alpha) = ||w - B\alpha||^2
\label{eq14}
\end{equation}

%The process of forward propagation is relatively simple, choose weights according to ALE and then use weights for calculation. We denote the function as follow:
For the forward pass, the network selects whether to binarize the weights of each layer according to the ALE and uses the weights to calculate the output of each layer $O$.
\begin{equation}
O =\begin{cases}\sum_{m=1}^{3}\alpha_mConv(B_m,A) & Binarization\\
            Conv(W,A) & else\end{cases}
\label{eq15}
\end{equation}
where $Conv()$ represents convolution function.

BSNNs are affected by binarized weight and binary input, so the backpropagation process must be reconsidered. We use the Dirac function to generate the spikes of SNNs.
Due to the non-differentiability of the Dirac function, the approximate gradient function is used instead of the derivative function in backpropagation\cite{neftci2019surrogate,wu2018spatio}, the approximate gradient function  is defined as follows:
\begin{equation}
h(u) = \frac{1}{a}sign(|u- V_{th}<\frac{a}{2}|)
\label{eq16}
\end{equation}
where $u$ represents the membrane voltage, $V_{th} $ represents the threshold, and $a$ is the parameter that determines the sharpness of the curve.

%We use the approximate gradient function in the chain rule of backpropagation to obtain a operation for processing binary inputs. Operation as follows:
Using the chain rule, the error gradient with the respect to the presynaptic weight W is
\begin{equation}
\frac{\partial{L}}{\partial{W}} = \frac{\partial{L}}{\partial{O}}\frac{\partial{O}}{\partial{W}} =  \frac{\partial{L}}{\partial{O}}(\frac{1}{a}sign(|u- V_{th}<\frac{a}{2}|))
\label{eq17}
\end{equation}
where $L$ is the loss function, and $sign$ is signum function.

Moreover, the binarization function of weight is also a typical step function, straight-through estimator (STE)\cite{bengio2013estimating} is usually used to solve this problem.
\begin{equation}
\frac{\partial{L}}{\partial{W}} \mathop{=}\limits^{STE} \frac{\partial{L}}{\partial{O}}\frac{\partial{O}}{\partial{B}}\frac{\partial{Htanh}}{\partial{W}} = \frac{\partial{L}}{\partial{O}}\frac{\partial{O}}{\partial{B}} =\frac{\partial{L}}{\partial{B}}
\label{eq18}
\end{equation}
where $O$ and $Htanh$ as the output tensor of a convolution and hard-tanh function respectively. 

\begin{figure*}[t]
 \centering 
 \includegraphics[width=12cm]{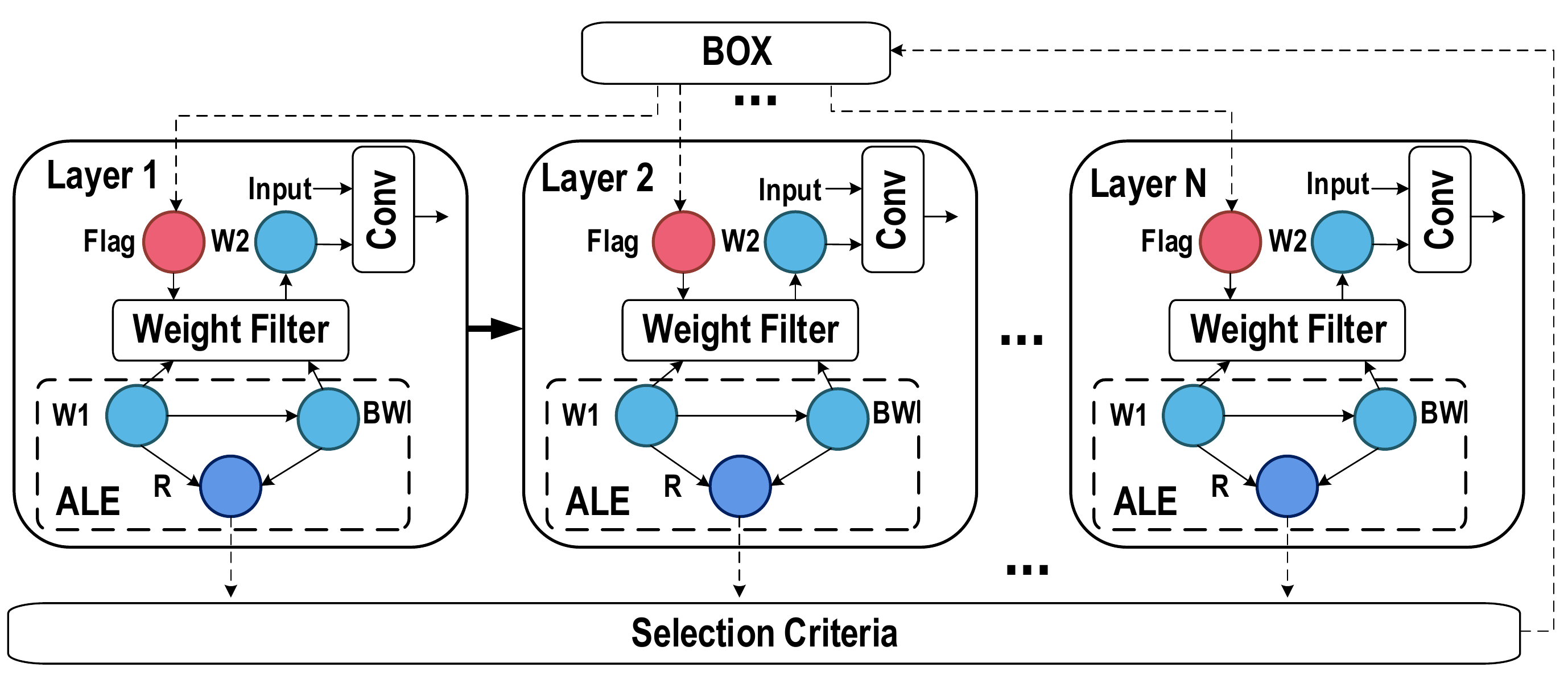}
 \caption{Network layers with ALE. The box is used to record the index of layers that need to be binarized. The flag is used to determine whether the binarized weights are used for convolution calculation. $W1$, $BW$ and $W2$ represent the original weight, the binarized weight, and the weights selected for convolution calculation, respectively. And $Conv$ is the convolution function.}
 \label{fig1}
 \end{figure*}

In Fig. \ref{fig1}, we show the network layer with ALE and its workflow.
Firstly, the network can use the Flag obtained from Box to determine whether this layer uses binarized weights.
Then convolve the selected weight with the input.
For the current time step, Box stores the selection result of the last time step, and these results will be used to select whether the binarized weight will be used.
ALE will recalculate the value of $R$ and update the selection results in the Box at the same time.
Next, the process for ALE to recalculate the value of $R$ is as follows. It calculates the binarized weight $BW$ according to the original weight $W1$, and then they work together to get $R$.
Finally, the selection result is dependent on the value of $R$ and the selection criteria, and the results are updated to the Box.
%The pipeline of our ALE is shown in Fig. \ref{fig1}.
%In addition, we provide pseudo-code for ALE's pipeline which is shown in algorithm\ref{alg:algorithm1}.\hl{**}

%\begin{algorithm}[tb]
%\caption{ALE's workflow\hl{**}}
%\label{alg:algorithm1}
%\textbf{Input}: $S_{in}$, TL, CL, K, W1.\\
%\textbf{Parameter}: $S_{in}$: Input Spiking Train, TL: index of the last layer, CL: index of current layer, K: number of classifications, W1: full precision weights, BW: binary weight, A: error estimation, M: coefficient to measure weight storage space, $N_{cout}$: number of output channels, box: used to save the index of the layer that needs to be binarized, Box: used to save the final selection result in Lbox.\\
%\textbf{Output}: Convolution result.
%\begin{algorithmic}[1] %[1] enables line numbers
%\STATE BW = CalBW(W1).   //Eq.\eqref{eq12}-(\eqref{eq14} 
%\STATE A = CalLoss(BW, W1, $N_{cout}$).   //Eq.\eqref{eq5}-\eqref{eq6}  
%\STATE M = CalM(W1,A).   //Eq.\eqref{eq7}-\eqref{eq8}
%\STATE R = CalR(TL, CL, A, l, K, M)   //Eq.\eqref{eq9}-\eqref{eq10}
%\STATE box $\leftarrow$ Judge whether CL needs binarization according to R
%\IF {CL == TL}
%\STATE Box = box.  
%\ENDIF
%\STATE Flag = .  
%\IF {Flag = 1}
%\STATE W2 = BW.
%\ELSE
%\STATE W2 = W1.
%\ENDIF
%\STATE output = Conv(W2).
%\end{algorithmic}
%\end{algorithm}

\subsection{GAP Layer}

 \begin{figure}[t]
 \centering 
 \includegraphics[width=8cm]{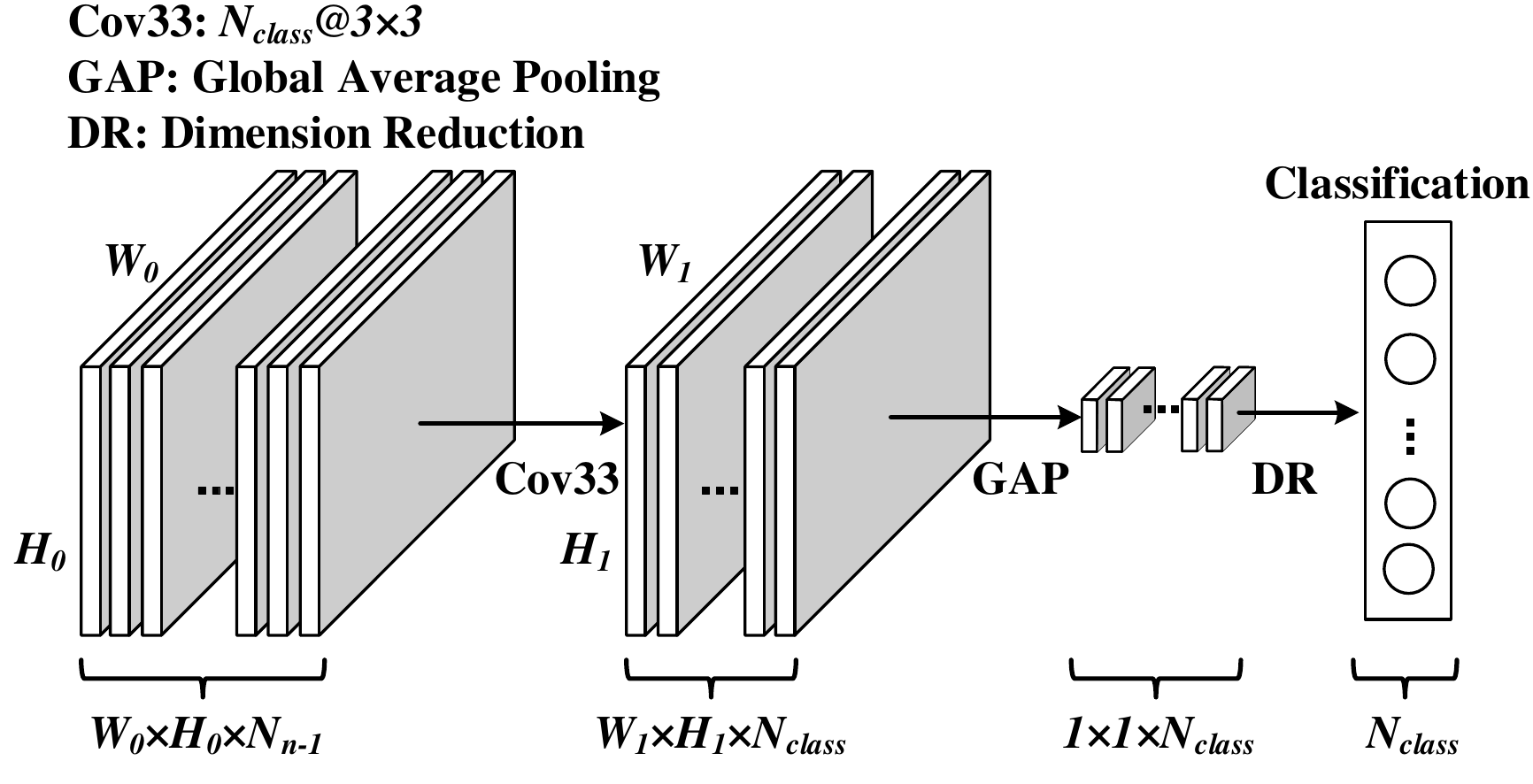}
 \caption{The overall structure of GAP Layer.}
 \label{fig2}
 \end{figure}
 
  \begin{figure*}[t]
 \centering %表示居中
 \includegraphics[width=14cm]{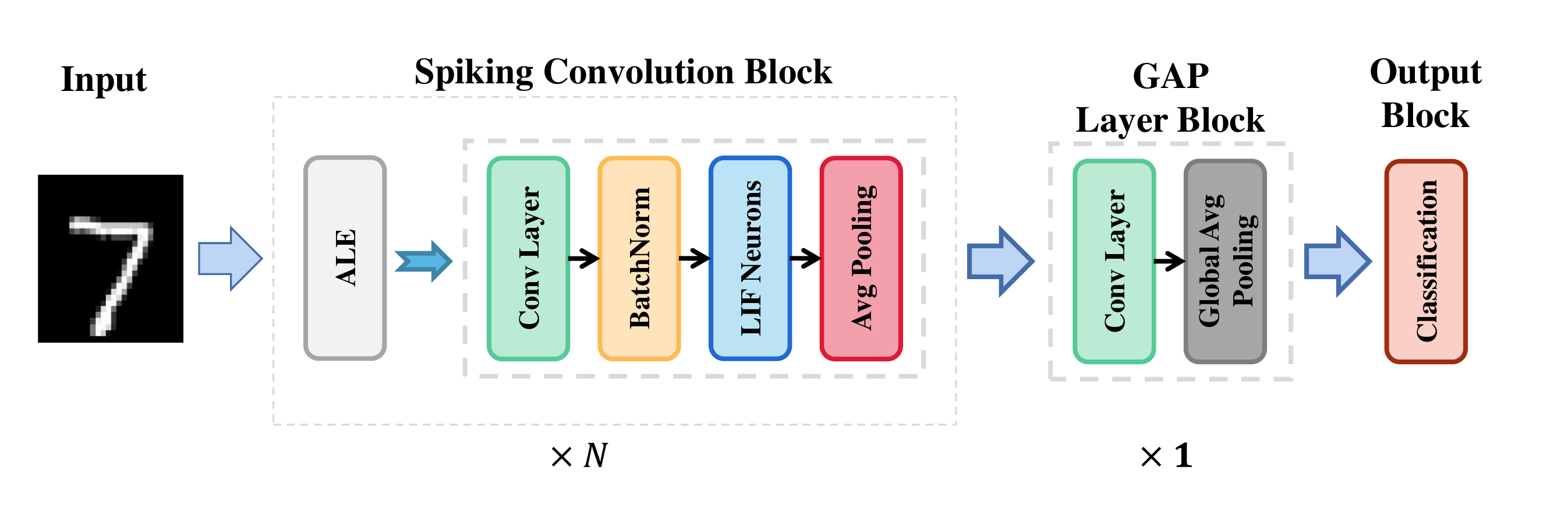} \caption{The overall structure of Adaptive Local Binary Spiking Neural Network.}
6 \label{fig3}
 \end{figure*}

%Because of the binary output of spiking neuron,it is extremely vulnerable to noise when the results of a single operation are directly used for classification.
%Therefore, it is generally believed that the output of the spiking network is the spiking frequency in a period of time, the spiking frequency indicates the degree of response to the category, hence causing generation of additional computational consumption.
%To address this problem, we learn from CNNs's global average pooling\cite{lin2013network} and use it in SNNs to reduce the time steps.

Because of the binary output of spiking neurons, it is extremely sensitive to noise when the results of a few time steps are directly used for classification.
Therefore, it is usually to use the spiking trains for a long period of time to indicate the degree of response to the category, which causes extra computational consumption.
To address this problem, we learn from CNN's global average pooling\cite{lin2013network} and apply it in SNNs to reduce the time steps.

GAP Layer consists of a convolutional layer and a global average pooling layer\cite{lin2013network}.
The convolution layer adjusts output channels to the number of classifications of the dataset.
The global average pooling layer is used to convert the feature map into a classification vector, which is directly related to the final classification result. 
The overall structure of the GAP Layer is shown in Fig. \ref{fig2}.
The number of output channels is adjusted to the number of dataset classes by convolution calculation, first.
Then a global average pooling is used to transform the spatial average of the feature maps from the last layer to the confidence of categories.
The obtained confidence is used as the probability of recognition.
Just as global average pooling plays a role in CNNs: it enforces correspondence between feature maps and categories and integrates global spatial information\cite{lin2013network}. We can naturally introduce these advantages into SNNs to alleviate the excessive cost of the time steps.
Therefore, SNNs can achieve competitive classification accuracy in few time steps or even one time step compared to existing state-of-the-art SNNs

\subsection{Adaptive Local Binary Spiking Neural Network}
The overall structure of the proposed Adaptive Local Binary Spiking Neural Network(ALBSNN) structure is illustrated in Fig. \ref{fig3}.
The network consists of $N$ end-to-end spiking convolution blocks, and a GAP layer block. 
The spiking convolution block consists of an ALE, a spiking convolution layer, a batch normalization layer, and an average pooling layer.
ALE decides whether the weight is binarized or not, and the spiking  convolution layer extracts the features of the image.
%It adaptively selects binary or full precision weights to extract information from images.
%The GAP Layer Block, it reduces the dimension of the upper block and maps the results to avoid adding extra computation and training burden. 
The GAP layer is used to alleviate the excessive cost of the time steps.
%The results processed by GAP Layer Block are used for classification. The overall structure of Adaptive Local Binary Spiking Neural Network is shown in figure \ref{fig3}.

\section{Experiments}
In this section, we evaluate our proposed Adaptive Local Binary Spiking Neural Network(ALBSNN) on three datasets Fashion-MNIST\cite{xiao2017fashion}, CIFAR-10, and CIFAR-100\cite{krizhevsky2009learning} datasets, which are the popular benchmarks for SNNs.
Fashion-MNIST is a fashion product image dataset with 10 classes, 70,000 grayscale images in the size of 28 × 28 are divided into 60,000 training data and 10000 test data.
CIFAR-10 and CIFAR-100 are composed of 3 channel RGB images of size 32×32 with 50,000 training images and 10,000 test images. CIFAR-10 has 10 classes, and CIFAR-100 has 100 classes, all images are divided equally by class.
 We verify the effectiveness of ALE and study the factors affecting ALE, first.
 Then we try four different selection criteria for binarization and compare them.
 Finally, we compared ALBSNN with several previously reported state-of-the-art results with the same or similar network.
%In this section, we evaluate two local binarization methods: (1) the first and last layers are not binarized (keep full precision), while other layers are binarized(the most common local binarization approach, which we call FLNB), (2) Accuracy Loss Estimator.In addition, we test the effects of different standards. Finally, we compare with related state-of-the-art networks from various aspects.

\subsection{Experimental Setup}

\begin{table}[t]
\centering
%z\resizebox{\columnwidth}{10mm}{
\begin{tabular}{ll}
\hline
  Dataset & Structure \\
  \hline
  Fashion-MNIST & \begin{tabular}[c]{@{}l@{}}16C3(Encoding)-16C3-AP2-64C3\\-64C3-AP2-256C3-1024C3-GAP\end{tabular}\\
  \makecell{CIFAR-10,\\CIFAR-100} & \begin{tabular}[c]{@{}l@{}}128C3(Encoding)-256C3-AP2\\-512C3-AP2-1024C3-512C3-GAP\end{tabular}\\
  \hline
  
\end{tabular}
%}
\caption{Network structures.}
\label{table1}
\end{table}

\begin{table}
\centering
\resizebox{\columnwidth}{20mm}{
\begin{tabular}{llllll} 
\toprule
  Parameter & Fashion-MNIST& CIFAR-10 & CIFAR-100 \\
  \midrule
  $V_{th}$ & 0.5 & 0.5 & 0.5\\
  $\tau$ & 0.25 & 0.25 & 0.25\\
  a & 1 & 1 & 1\\
  %epoch &  20 & 200 & 20\\
  learning rate & 0.001 & 0.001 & 0.001\\
  batch size & 16 & 16 & 64\\
  time step & 1 & 1 & 1\\
  optimizer & Adam & Adam & Adam\\
  criterion & MSE & MSE& Cross-Entropy \\
  %Structure &\makecell{16C3-16C3\\-AP2-64C3\\-64C3-AP2\\-256C3-\\1024C3-GAP}&\makecell{128C3-256C3\\-AP2-512C3\\-AP2-1024C3\\-512C3-GAP}&\makecell{128C3-256C3\\-AP2-512C3\\-AP2-1024C3\\-512C3-GAP}\\
  \bottomrule
  \end{tabular}
 }
    \caption{Parameters setting.}
  \label{table2}
\end{table}

%We evaluate on Fashion-MNIST\cite{xiao2017fashion}, CIFAR-10 and CIFAR-100\cite{krizhevsky2009learning} datasets,which are popular benchmark for SNNs. Fashion-MNIST is a fashion product image dataset with 10 classes, 70,000 grayscale images in the size of 28 × 28 are divided into 60,000 training data and 10000 test data.
%CIFAR-10 and CIFAR-100 are composed of 3 channel RGB images of size 32×32 with 50,000 training images and 10,000 test images. CIFAR-10 has 10 classes,and CIFAR-100 has 100 classes, all images are divided equally by class.

All reported experiments below are conducted on an NVIDIA Tesla V100 GPU.
The implementation of our proposed ALBSNN is on the Pytorch framework\cite{PyTorch}.
Only one time step is used to demonstrate the advantage of our proposed ALBSNN on ultra low-latency.
Adam is applied as the optimizer\cite{Adam}.
The results shown in this paper refer to the average results obtained by repeating five times.

%To improve the robustness and generalization ability of the model,
In this paper, we apply several data augmentation during training processing as follows: (1) padding the original figure, and the padding size is 4, (2) crop pictures with a size of 32 pixels randomly, (3) flip the image horizontally with half probability, (4) normalized image,the standard deviation is 0.5.
For the testing process, only normalization is applied\cite{shorten2019survey}.

%\begin{table}[t]
%\centering
%\begin{tabular}{ll}
%\toprule
%  Type & Structure \\
%  \midrule
%  Structure-1 & \begin{tabular}[c]{@{}l@{}}128C3(Encoding)-256C3-AP2-512C3\\-AP2-1024C3-512C3-GAP\end{tabular}\\
%  Structure-2 & \begin{tabular}[c]{@{}l@{}}16C3(Encoding)-32C3-AP2-64C3\\-128C3-AP2-64C3-32C3-GAP\end{tabular}\\
%  Structure-3 & \begin{tabular}[c]{@{}l@{}}16C3(Encoding)-16C3-AP2-64C3\\-64C3-AP2-256C3-1024C3-GAP\end{tabular}\\
%  Structure-4 & \begin{tabular}[c]{@{}l@{}}16C3(Encoding)-32C3-AP2-512C3\\-AP2-512C3-1024C3-GAP\end{tabular} \\
%  \bottomrule
%\end{tabular}
%}
%\caption{Network structures used for accuracy evaluation.}
%\label{table1}
%\end{table}

We use iterative LIF model and approximate gradient for network training.
The first convolutional layer acts as an encoding layer and network structures for Fashion-MNIST, CIFAR-10, and CIFAR-100 datasets are shown in Table \ref{table1}.
Between the convolution calculation and the activation function, batch-normalization(BN)\cite{ioffe2015batch} is applied.
All convolution operations used in the experiment are based on the operations provided by Pytorch.
%Table \ref{table1} shows the neural network structure of training
The hyperparameters of netowkrs we used in our experiments are shown in Table \ref{table2}.
The learning rate uses the cosineanealing strategy\cite{loshchilov2016sgdr}.
Unless otherwise specified, the testing accuracy of Fashion-MNIST and CIFAR-10 are reported after training 20 epochs in our experiments.
And for CIFAR-100, 200 epochs are applied for training.

%The hyperparameters we used in our experiments are shown in Table \ref{table2}. The learning rate uses the cosineanealing strategy\cite{loshchilov2016sgdr}, we use the default initialization of PyTorch, experiment with one GPU at the same time.

\subsection{Effectiveness of ALE}
%We evaluate ALE and FLNB on Fashion-MNIST and CIFAR-10 datasets, respectively. 
%In order to observe and compare, set ALE to select two layers to maintain full precision (select the two layers with the largest value). 
%Table \ref{table3} shows the accuracy and binarization of the two methods on the same network structure. The above results show that the selection results of ALE are better than FLNB. 
To validate the effectiveness of ALE, We compared ALBSNN with SNN with full precision weights (FPSNN), Binarized SNN (BSNN), and BSNN whose first layer and last layer are non-binarized(FLNBSNN) on Fashion-MNIST, CIFAR-10, and CIFAR-100 datasets, respectively. 
For the fairness of comparison, ALBSNN is designed to select two layers to maintain full precision. 
%Table \ref{table3} shows the accuracy of the two methods on the same network structure.
%As the table shows that the ALBSNN with ALE is better than FLNBSNN.
%The results show that although two layers are selected non-binarized for both networks, ALBSNN achieves better results in accuracy because the ALE block can help ALBSNN select more suitable layers based on the network structure and dataset.
Table \ref{table4} shows the accuracy of four different methods.
%We obtain full precision (FP) and full binarization (FB) results by STBP\cite{wu2018spatio} and ABC-NET\cite{lin2017towards}.
We obtain FPSNN and BSNN results by STBP\cite{wu2018spatio} and ABC-NET\cite{lin2017towards}.
Compared with FPSNN, BSNN, FLNBSNN, and ALBSNN will drop a some accuracy due to binarization.
Among three binarized SNNs, ALBSNN achieves better results in accuracy because the ALE block can help ALBSNN select more suitable layers based on the network structure and dataset.
For Fashion-MNIST and CIFAR-10 dataset, compared with FPSNN, ALBSNN only drops 0.20$\%$, and 0.52$\%$ accuracy, respectively.
The reason that ALBSNN drops 3.49$\%$ accuracy for the CIFAR-100 dataset is that ALBSNN is limited to select two layer to maintain full precision.
In the section about selection criteria below, we will show the improvement of ALBSNN by different selection criteria.

There is an interesting phenomenon that FLNBSNN and ALBSNN both select the first and sixth layer as non-binarized layers for the CIFAR-10 and CIFAR-100 datasets, but ALBSNN obtains a better accuracy.
After further study, we find ALBSNN does not always select the first and sixth layers as non-binarized layers.
For example, ALBSNN only selects the sixth layer as non-binarized layers at the first iteration of training for CIFAR-10.
In a further experiment, we force the first layer of the first forward in FLBSNN to be binarized so that it has the same effect as ALBSNN, other conditions remain unchanged. 
We can observe that the recognition accuracy of CIFAR-10 is improved from 89.17$\%$ to 90.02$\%$.
%It is worth noting that the number of layers that keep full precision in ALE is not consistent with FLNB,although both choose the same two layers to keep full precision. 
%Such as FLNB and ALE of Fashion-MNIST in Table \ref{table4},in FLNB, both the first and last layers maintain full precision 87500 times. 
%In ALE, the first and last layers maintain full precision of 87499 and 87500 times, respectively.
%In a further experiment, we force the first layer of the first forward in FLNB to be binarized so that it has the same effect as ALE, other conditions remain unchanged. 
%We can observe that the recognition accuracy of Fashion-MNIST is improved from 92.21 to 92.24, CIFAR-10 is improved from 89.17 to 90.02.
Therefore, we believe that the binarized weight used in the initial input can be regarded as beneficial noise, it can improve the robustness of the network.

\begin{table}
\centering
%\resizebox{\columnwidth}{32mm}{
\begin{tabular}{llllll} 
\toprule
   Dataset  & Method & \makecell{Full\\precision\\layer} & Acc($\%$) \\
  \midrule
\multirow{4}{*}{Fashion-MNIST~}  &BSNN &  - & 92.38 \\
&FLNBSNN &   1,6 & 92.92 \\
&ALBSNN & 1,2 & 93.10 \\
&FPSNN & all & 93.30 \\
\multirow{4}{*}{CIFAR-10~} & BSNN  & - & 88.53 \\
&FLNBSNN   &  1,6 & 89.17 \\
&ALBSNN   & 1,6 & 90.12 \\
&FPSNN   & all & 90.64 \\
\multirow{4}{*}{CIFAR-100~} & BSNN   & - & 57.68 \\
&FLNBSNN   &  1,6 & 63.31 \\
&ALBSNN   & 1,6 & 63.54 \\
&FPSNN   & all & 67.03 \\
  \bottomrule
  \end{tabular}
%  }
    \caption{Accuracy of different methods.}
  \label{table4}
\end{table}

\subsection{Rethink about local binarization}
Compared the selection results binarized layer for Fashion-MNIST, CIIFAR-10, and CIFAR-100, we find these selection result is related to the complexity of dataset and the network structure.
As shown in Table \ref{table4}, ALBSNN chooses the same layers as FLNSNN to keep full precision when the dataset  is CIFAR-10 or CIFAR-100.
Compared with the network structure we used, We find that if final output channel is relatively small and the size of weights between adjacent network layers is relatively large, ALBSNN may obtain a better binarization scheme by ALE.
%We find that when final output channel is relatively small (with few categories of dataset) and the weight gap between the front layer and the back layer is relatively large, a better selection than FLNB is more likely to appear.
However, if the size of weights in the network increases or decreases gradually, FLNBSNN is a good solution. 
As the weights of common networks generally conform to the rule of flat change layer by layer, therefore, the selection of ALE tends to be similar to FLNB. 
Of course, if the non-binarized layers are not limited to two, ALE still can obtain a better binarization scheme by evaluating the error caused by the binarized weights.
To sum up, the selection result of ALE is mainly related to the complexity of the dataset and structure of the neural network

\subsection{Impact of selection criteria}

\begin{table}
\centering
%\resizebox{\columnwidth}{30mm}{
\begin{tabular}{llll} 
\toprule
   Dataset & \makecell{Selection\\criteria}  & \makecell{Full\\precision layer} & Acc($\%$) \\
  \midrule
  \multirow{4}{*}{Fashion-MNIST~}  &SC1   & 1 & 92.81 \\
  &SC2   & 1,6 & 93.10\\
  &SC3  & 1,2,7 & 93.26 \\
  &SC4   & 1,3,7 & 93.21  \\
   \multirow{4}{*}{CIFAR-10~}&SC1    & 1 & 89.49\\
  &SC2  & 1,6 & 90.12\\
  &SC3    & 1,2,6 & 90.12\\
  &SC4   & 1,5,6 & 90.15\\
   \multirow{4}{*}{CIFAR-100~}&SC1    & 1 & 60.86\\
  &SC2   & 1,6 &63.54\\
  &SC3    & 1,2,6 & 64.23\\
  &SC4   & 1,5,6 & 64.54\\
  \bottomrule
  \end{tabular}
 % }
  \caption{Accuracy of different selection criteria.}
  \label{table5}
\end{table}

In the previous section, in order to make a fair comparison with FLNBSNN, we select the two layers with the largest value $R$ as full precision layers.
In this section, we choose four different selection criteria  SC1, SC2, SC3, and SC4, to show the impact of the selection criteria on the accuracy of ALBSNN.
SC1 applies the mean value $R$ of all layers as the baseline. When the value $R$ of a layer is greater than the mean value, this layer is selected as the full precision layer.
SC2 uses the $R$ of the last layer as the baseline. 
If the $R$ of a layer is greater than the baseline, and the layer is non-binarized.
For SC3, the first and last layers are selected as full precision layers, and the mean of $R$ of the other layers is set as the baseline, if the $R$ of other layers exceeds the baseline, the layer is selected as the full precision layer.
For SC4, the first and last layers are selected as full precision layers, and the layer closest to the average value of $R$ excluding these two layers is also regarded as the full precision layer.

As the Table \ref{table5} is shown, different binarization scheme is obtained based on the network structure and dataset by ALE with the different selection criteria.
For the CIFAR-100 dataset, the accuracy can be improved from 63.54$\%$ to 64.54$\%$ by only adding one non-binarized layer.
In practice, we can choose the appropriate selection criteria according to the requirements of accuracy and weight storage space.

\subsection{Compared with other methods}

\begin{table*}
\centering
\begin{tabular}{lll} 
\toprule
   Dataset&Method & Structure \\
  \midrule
  \multirow{4}{*}{\makecell{Fashion\\-MNIST~}}  &BS4NN   & 600FC-600FC-10 \\
  &SSTiDi-BP  & 20C5-MP2-40C5-MP2-1000FC-10\\
  &ALBSNN  &  20C3-MP2-40C3-MP2-1000C3-10 \\
  \multirow{4}{*}{CIFAR-10~}  &Roy-SVGG10  & 128C3$\times$2-MP-256C3$\times$2-MP2-512C3$\times$2-MP2-1024FC-1024FC-10 \\
  &Wang-SVGG10 & 128C3$\times$2-MP2-256C3$\times$2-MP2-512C3$\times$2-MP2-1024FC-1024FC-10\\
  &ALBSNN  & 128C3-256C3-AP2-512C3-AP2-1024C3-512C3-10 \\
  \multirow{4}{*}{CIFAR-100~}  &Roy-SVGG100  & 64C3$\times$2-MP2-128C3$\times$2-MP2-256C3$\times$3-MP2-(512C3$\times$3-MP2)$\times$2-4096FC-4096FC-100 \\
  &Wang-SVGG100  & 128C3$\times$2-MP2-256C3$\times$2-MP2-512C3-512C3-MP2-1024FC-1024FC-512FC-100\\
  &ALBSNN & 128C3-256C3-AP2-512C3-AP2-1024C3-512C3-100\\
  \bottomrule
  \end{tabular}
%\begin{tablenotes}
%%\item[1] 
%128C3$\times$2 represents 2 convolution block, each convolution block with 128 3 $\times$ 3 filters.
%AP2 represents average pooling layer with 2 $\times$ 2 filters.
%MP2 represents max pooling layer with 2 $\times$ 2 filters.
%%DP denotes dropout layer and 
%600FC means a fully connected layer that consists of 600 neurons.
%\end{tablenotes}
  \caption{Structure of the network with different methods.}
  \label{table6}
\end{table*}

\begin{table*}
\centering
\begin{tabular}{lllllll} 
\toprule
   Dataset&Method & Learning & Epoch & Timestep & \makecell{Weight storage space\\(Normalized)} & Acc($\%$) \\
  \midrule
  \multirow{4}{*}{Fashion-MNIST~}  &BS4NN   & Temporal backpropagation & 500 & 100 & 1.85 & 87.50 \\
  &SSTiDi-BP  & Temporal backpropagation & - & 100 & 3.09 & 92.00 \\
  &ALBSNN  & STBP & 20 & 1 & 1 &  91.83\\
  \multirow{4}{*}{CIFAR-10~}  &Roy-SVGG10  & ANN2SNN & 150 & - & 1.26 &  88.27 \\
  &Wang-SVGG10  & ANN2SNN & 500  & 100 & 1.26 &  90.19\\
  &ALBSNN  & STBP & 50 & 1 & 1 & 91.63 \\
  \multirow{4}{*}{CIFAR-100~}  &Roy-SVGG100  & ANN2SNN & 400 & - & 2.76 & 54.44 \\
  &Wang-SVGG100  & ANN2SNN & 500  & 300 & 1.18 & 62.02\\
  &ALBSNN  & STBP & 200 & 1 & 1 & 63.54\\
  \bottomrule
  \end{tabular}
  \caption{Compare with different methods.}
  \label{table7}
\end{table*}

In this section, we compare our proposed ALBSNN with several previously reported state-of-the-art method with the same or similar network.
%Table \ref{table7} summarizes the characteristics and recognition accuracies of recent BSNNs on these three datasets. 
For the Fashion-MNIST, BS4NN\cite{kheradpisheh2022bs4nn} is trained with a simple fully connected network, and \cite{MIRSADEGHI2021131} uses a higher-performance convolutional network for recognition(We denote this network by SSTiDi-BP). Both networks use temporal backpropagation for learning. 
For a fair comparison, we replace the fully connected layer with the GAP Layer and build an ALBSNN based on a similar network structure for discussion.
For CIFAR-10 and CIFAR-100 datasets, the network structures used by \cite{roy2019scaling} and \cite{wang2020deep} are both modified VGG network\cite{simonyan2014very}, we use Roy-SVGG10 and Wang-SVGG10 to denote these two networks, respectively.
%They both use a more common approach to dealing with complex datasets: convert ANN to SNN.
They do not train directly the SNN but rather use the method of ANN-to-SNN conversion.
For CIFAR-10 and CIFAR-100 datasets, we also build an ALBSNN based on a similar network structure to make the comparison.
%However, the network structure (stru-1) used to identify the CIFAR-10 and CIFAR-100 datasets in this paper is similar to them, so we use the results of stru-1 directly for comparison and analysis.
%The network structure is shown in Table \ref{table6}, $AP$ denotes average pooling, $MP$ denotes max pooling, they all use a pooling kernel of size 2 for pooling with stride 2. $128C$ represents a convolutional layer with an output channel of 128 and a convolution kernel size of 3 (except that SSTiDi-BP in Fashion-MNIST uses a convolution kernel of size 5, the size of other convolution kernels all default to 3).

Table. \ref{table6} list the network structure of different methods, in which 128C3$\times$2 represents 2 convolution block, each convolution block with 128 3 $\times$ 3 filters, AP2 represents average pooling layer with 2 $\times$ 2 filters, MP2 represents max pooling layer with 2 $\times$ 2 filters, and 600FC means a fully connected layer that consists of 600 neurons.
Table. \ref{table7} shows the results of different methods.
The weight storage space is normalized with respect to the baseline(ALBSNN).
For Fashion-MNIST datasets, our recognition accuracy is on the same level as state-of-the-art networks, but we use less training time and save more than 45$\%$ storage resources.
%(in Table \ref{table6}, we set the network weight size obtained by the method in this paper as the comparison standard, and the value is 1).
For CIFAR-10 and CIFAR-100 datasets, our proposed ALBSNN obtained 91.63$\%$ and 63.54$\%$ accuracy, respectively.
Compared with Wang-SVGG10, our proposed ALBSNN achieves 1.44$\%$ and 1.52$\%$ average testing accuracy improvement with only one time steps and fewer epochs.
For the weight storage space, our proposed ALBSNN can obtain more than 20$\%$ and 15$\%$ reduction on the CIFAR-10 and CIFAR-100 datasets, respectively.
%For CIFAR-100, we can not only save much more storage resources and training time, but also achieve competitive classification accuracy compared to existing state-of-the-art BSNNs.

\section{Conclusion}
This paper proposes a construction method of Ultra-low Latency Adaptive Local Binary Spiking Neural Network with Accuracy Loss Estimator, which balances the pros and cons between full precision weights and binarized weights by choosing binarized or non-binarized weights adaptively.
Our proposed network satisfies the requirement of network quantization while keeping high recognition accuracy.
At the same time, we find the problem of long training time for BSNNs.
Therefore, we propose the GAP Layer, in which a convolution layer is used to replace the fully connected layer, and a global average pooling layer is used to solve the binary output problem of SNN.
Because of the binary output, SNN usually needs to run multiple time steps to get reasonable results.
Experiments on Fashion-MNIST, CIFAR-10, and CIFAR-100 show that our method not only saves more storage resources and training time, but also achieves competitive classification accuracy compared with existing state-of-the-art BSNNs

% Use \bibliography{yourbibfile} instead or the References section will not appear in your paper
\bibliography{cite.bib}

\end{document}